\documentclass{article}

\usepackage{microtype}
\usepackage{graphicx}
\usepackage{subcaption}
\usepackage{booktabs} 
\usepackage{tabularx}
\newcolumntype{Y}{>{\centering\arraybackslash}X}

\usepackage{array}

\usepackage[most]{tcolorbox}
\tcbuselibrary{listings,breakable}
\usepackage{listings}
\newtcblisting{promptbox}{
  breakable,
  colback=gray!5,
  colframe=gray!50,
  boxrule=0.5pt,
  arc=2pt,
  left=6pt,right=6pt,top=1pt,bottom=1pt,
  before skip=6pt,
  after skip=6pt,
  listing only,
  listing options={
    basicstyle=\ttfamily\small,
    breaklines=true,
    columns=fullflexible
  }
}

\PassOptionsToPackage{hyphens}{url}\usepackage{hyperref}

\usepackage[accepted]{icml2026}

\usepackage{amsmath}
\usepackage{amssymb}
\usepackage{mathtools}
\usepackage{amsthm}

\usepackage[capitalize,noabbrev]{cleveref}

\theoremstyle{plain}

\theoremstyle{definition}

\theoremstyle{remark}

\usepackage[textsize=tiny]{todonotes}

\icmltitlerunning{Inverting Brains with SBI}

\newcommand{\x}{\mathbf{x}}

\newcommand{\z}{\mathbf{z}}
\newcommand{\s}{\mathbf{s}}
\newcommand{\X}{\mathbf{X}}

\newcommand{\stim}{\mathbf{S}}
\newcommand{\thetab}{\boldsymbol{\theta}}

\newcommand{\phib}{\boldsymbol{\phi}}
\newcommand{\epsilonb}{\boldsymbol{\epsilon}}

\begin{document}

\twocolumn[
  \icmltitle{Inverting Foundation Models of Brain Function with Simulation-Based Inference}



  \icmlsetsymbol{equal}{*}

  \begin{icmlauthorlist}
    \icmlauthor{Niels Bracher}{sch}
    \icmlauthor{Xavier Intes}{sch}
    \icmlauthor{Stefan T.~Radev}{sch}
  \end{icmlauthorlist}

  \icmlaffiliation{sch}{Center for Modeling, Simulation, \& Imaging in Medicine, Rensselaer Polytechnic Institute, USA}

  \icmlcorrespondingauthor{Niels Bracher}{brachn@rpi.edu}
  \icmlcorrespondingauthor{Stefan T.~Radev}{radevs@rpi.edu}

  \icmlkeywords{Machine Learning, ICML}

  \vskip 0.3in
]



\printAffiliationsAndNotice{}  

\begin{abstract}
Foundation models of brain activity promise a new frontier for \textit{in silico} neuroscience by emulating neural responses to complex stimuli across tasks and modalities. A natural next step is to ask whether these models can also be used in reverse. Can we recover a stimulus or its properties from synthetic brain activity?
We study this question in a proof-of-concept setting using TRIBEv2. We pair the brain emulator with large language models (LLMs) that generate news headlines from linguistic parameters such as valence, arousal, and dominance. We then use simulation-based inference to learn a probabilistic mapping from brain maps to latent stimulus parameters.
Our results show that these parameters can be recovered from predicted brain maps, demonstrating that the emulator's synthetic neural encodings preserve information about the controlled stimulus dimensions.
They also show that LLMs can serve as controllable stimulus generators for simulated experiments. Together, these findings provide a step toward decoding and inverse design with foundation brain models.
\end{abstract}

\section{Introduction}

A central goal in neuroscience is to understand and predict how properties of our environment are encoded in neural activity patterns \citep{van2017primer}.~Motivated by the rise of large language models, there is growing interest in building \textit{foundation models of brain function} that can predict neural activity across tasks and modalities \citep{dAscoli2026TribeV2Meta, wang2025foundation, dong2024brain, caro2023brainlm}. If faithful, such emulators realize the dream of \textit{in silico} neuroscience. For example, recent results from TRIBEv2 \citep{dAscoli2026TribeV2Meta} indicate that multimodal foundation models can reproduce known visual and linguistic effects, thus rapidly narrowing the simulation gap.

A natural next question is whether these foundation models can also be used in reverse: which inputs are consistent with a desired pattern of brain activity? The answer to this \textit{decoding} question is essential for designing informative stimuli \citep{ponce2019evolving, bashivan2019neural} and exploring new ways to probe neural representations via model-driven synthesis \citep{walker2019inception, yamins2016using}.
Furthermore, inverting foundation emulators can enable \textit{inverse design} of stimuli tailored to evoke specific brain states or differentiate between competing mechanistic hypotheses \citep{kar2019evidence}.

At the same time, inverting foundation models is challenging for both computational and modeling reasons. On the one hand, the forward mapping from stimuli to brain responses is high-dimensional, nonlinear, and opaque, so the inverse problem is typically ambiguous. On the other hand, systematically testing inversion requires a \textit{parameterized stimulus generator}: a model capable of synthesizing stimuli from interpretable latent variables \citep{kar2019evidence}.

\begin{figure*}[t]
    \centering
    \includegraphics[width=0.99\textwidth]{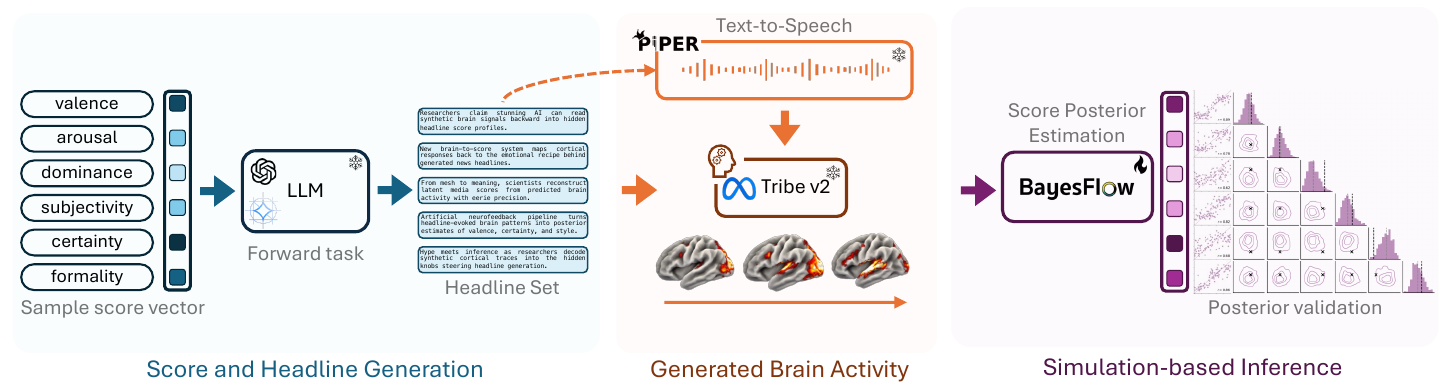}
    \caption{\textit{Overview of our simulation-based inference (SBI) pipeline for decoding from foundation brain models}. LLMs generate text stimuli from linguistic scores (latent parameters), which are subject to cycle consistency and converted to audio with PIPER. The stimuli are fed into the multimodal TRIBEv2 foundation model to generate time-resolved synthetic brain maps. The stimulus generator and brain emulator together form a composite simulator, which we invert probabilistically using SBI to recover the stimulus properties from brain activations. The pipeline can be used to validate neural encoding, perform inverse stimulus design, and enable model-driven synthesis.}
    \label{fig:overview}
\end{figure*}

In this proof-of-concept study, we make a step towards addressing these challenges by framing the decoding of stimulus features from foundation brain models as a simulation-based inference (SBI) problem (see~\autoref{fig:overview}). Our contributions are twofold:
\begin{enumerate}
\itemsep0em
    \item We show that frontier LLMs can be inserted into the SBI loop as stimulus designers conditioned on explicit linguistic parameters encoding emotional and structural properties.
    \item Using SBI with a relatively small training budget, we then show that brain encodings from TRIBEv2 contain sufficient information to recover the linguistic properties of the underlying stimulus.
\end{enumerate}

\section{Background}

\paragraph{Neural emulators}
In this work, a \emph{neural emulator} $\mathcal{M}(\cdot ; \phib)$ is a parameterized predictive model with pretrained parameters $\phib$ that maps multimodal representations of an external stimulus $\stim_t$ to a spatial brain activation map $\x_t \in \mathbb{R}^{V}$ of size $V$ at time $t$, as in models such as TRIBEv2 \citep{dAscoli2026TribeV2Meta}. We write $\stim_t = \bigl(\s_1(t), \dots, \s_m(t)\bigr)$ for the stimulus decomposed into $m$ modalities.

\paragraph{Stimulus generators and synthetic data}
In many experiments, the set of experimental stimuli $\mathcal{S} = \{\stim_1,\dots,\stim_T\}$ is not generated from an explicit parameterized model, but is instead selected heuristically or based on prior work. For example, an experimenter may choose a small set of hand-picked and hand-labeled images or sentences to probe semantic processing, without accessing a stimulus generator.

However, if we have access to a \textit{parameterized stimulus generator}, then we could simulate complete synthetic neuroscience experiments. In that case, latent parameters $\thetab$ define a distribution over stimuli, $\mathcal{S} \sim p(\mathcal{S} \mid \thetab)$, which can then be passed through the emulator to generate synthetic brain responses $\X = (\x_1,\dots,\x_T)$. Repeating this process yields synthetic triples $(\X_n, \mathcal{S}_n, \thetab_n)$ that can be used for downstream training of inverse models.

\paragraph{Neural simulation-based inference}
Neural simulation-based inference provides a flexible framework for Bayesian inversion \citep{Cranmer2020-SBI, Deistler2025-SBI, arruda2025diffusion}. In neural SBI, simulations from a forward model are typically used to train a neural density estimator (i.e., a conditional generative network) that approximates a probabilistic solution to the inverse problem.

In our setting, given a prior $p(\thetab)$ over a ``stimulus-generating'' distribution $p(\mathcal{S}\mid\thetab)$ and a neural emulator $\mathcal{M}$, we can run synthetic experiments by simulating:
\begin{equation}
    \thetab_n \sim p(\thetab), \quad \mathcal{S}_n \sim p(\mathcal{S}\mid\thetab_n), \quad \X_n = \mathcal{M}(\mathcal{S}_n;\phib).
\end{equation}
This simulation yields the triples $(\X_n,\mathcal{S}_n,\thetab_n)$ that serve as training data for the neural estimator.

In principle, one could learn a joint model $q(\mathcal{S},\thetab \mid \X)$ to decode both the stimuli and their latent properties $\thetab$ from brain activation. In this work, however, we focus on the inverse design setting and therefore learn a model $q(\thetab \mid \X)$ which maps observed brain activation to the parameters governing the stimulus generator (see~\autoref{fig:overview}). 

\section{Method}

\subsection{Data generation and validation}\label{sec:method-data-cc}

\paragraph{Frontier LLMs as stimulus generators}

To instantiate the abstract stimulus-generating distribution $p(\mathcal{S}\mid\thetab)$, we use frontier large language models (LLMs) as controllable generators of textual stimuli. For each sample $n$, we first draw a six-dimensional linguistic parameter vector $\thetab_n \in [0,1]^6$ from a uniform prior $p(\thetab)$, with components corresponding to \textit{valence}, \textit{arousal}, \textit{dominance}, \textit{subjectivity}, \textit{certainty}, and \textit{formality}.

\begin{figure*}[t]
    \centering
    \includegraphics[width=1\linewidth]{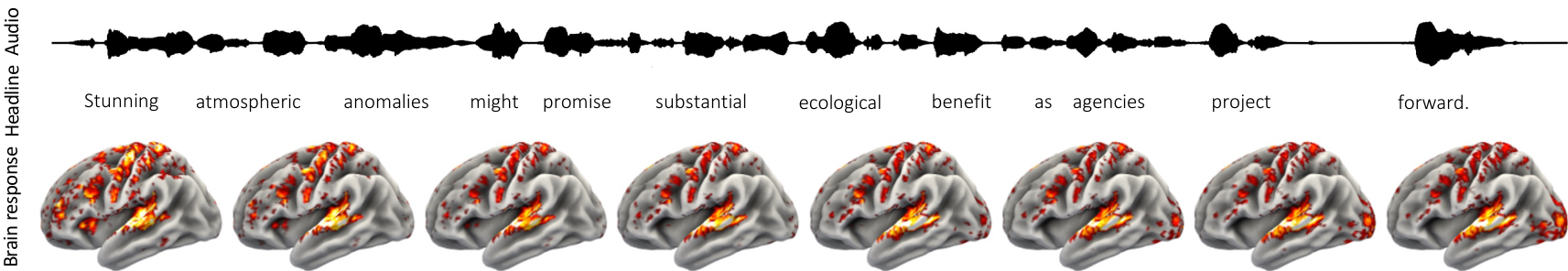}
    \caption{\textit{Example activation sequence predicted by TRIBEv2 for one synthetic headline.} 
    The panels show emulator-generated cortical response maps aligned to successive stimulus segments, with prominent predicted activity in lateral temporal cortex, inferior frontal cortex, and dorsal frontoparietal regions. The sequence is shown as a qualitative visualization of the emulator output rather than as evidence for a mechanistic interpretation of human neural activity.}
    \label{fig:headline-brain-response}
\end{figure*}

We choose these linguistic dimensions because they span two complementary layers of variation \citep{pavlick2016empirical, warriner2013norms, wilson2005opinionfinder}:
\begin{itemize}
    \item \textit{Affect} -- how language encodes emotional sentiment, tone, and charge;
    \item \textit{Stance} -- how language encodes the speaker's attitude, commitment, and register.
\end{itemize}
Each prompt is additionally conditioned on a topic variable $c_n$; in our experiments, $c_n$ ranges over ten comparatively neutral topics, such as \textit{sports}, \textit{science}, and \textit{weather}.

Given $(\thetab_n, c_n)$, the LLM generates a fixed-format set of five news headlines,
\begin{equation}\label{eq:forward}
    \mathcal{S}^{\mathrm{text}}_n \sim \mathrm{LLM}
    \left(\mathrm{User}\{\thetab_n, c_n\}, \mathrm{System} \right),
\end{equation}
implemented through a two-part prompting scheme: \textbf{(1)} a system prompt specifies how the latent dimensions map to linguistic properties and enforces the output format; and \textbf{(2)} a user prompt provides the topic variable $c_n$ and desired text parameters $\thetab_n$. Verbatim prompt templates are provided in \autoref{app:data-generation}.

\paragraph{Text-to-speech conversion} Since TRIBEv2 predicts temporally resolved brain activity $\x_1,\dots,\x_T$, the generated text block $\mathcal{S}^{\mathrm{text}}_n$ alone is not yet a sufficient stimulus representation. We therefore synthesize corresponding speech, denoted by $\mathcal{S}^{\mathrm{speech}}_n$, using a text-to-speech (TTS) generator \citep{pipertts2026}, and combine both modalities into a temporally aligned stimulus sequence
\begin{equation}
    \mathcal{S}_n = \{\stim_{n,t}\}_{t=1}^{T_n},
    \qquad
    \stim_{n,t} = \bigl(\s^{\mathrm{text}}_n(t), \s^{\mathrm{speech}}_n(t)\bigr),
\end{equation}
which provides the multimodal input required by TRIBEv2. An example of a generated headline with corresponding speech and brain activity is shown in \autoref{fig:headline-brain-response}.

\paragraph{LLM validation with cycle consistency}

We additionally assess whether the headline score vector $\thetab$ remains recoverable from the generated headline block alone, and thus whether the LLM generator heeds $\thetab$. Given a generated headline set $\mathcal{S}^{\mathrm{text}}_n$, we apply an \textit{inverse LLM evaluator},
\begin{equation}
    \hat{\thetab}^{\mathrm{text}}_n \sim \mathrm{LLM} \left(\mathrm{User}^{-1}\{\mathcal{S}^{\mathrm{text}}_n\}, \mathrm{System}^{-1} \right),
\end{equation}
which estimates one shared score per dimension for the headline set as a whole. The recovered scores \smash{$\hat{\thetab}^{\mathrm{text}}_n$} are then compared with the original latent parameters $\thetab_n$ used for generating the headlines (Eq.~\ref{eq:forward}).
The exact system prompt and a sample-specific user prompt used for this inverse evaluation are provided in the \autoref{app:data-generation}.

\subsection{Simulation-based inference setup}

\paragraph{Composite neural estimator}
Our brain-to-score amortized estimator consists of two jointly trained components: a brain summary backbone and a generative inference network \citep{radev2020bayesflow}.~The summary backbone transforms high-dimensional neural response sequences into a compact representation; the inference network maps this representation to a posterior distribution over the latent headline parameters.

\paragraph{Summary backbone} The summary backbone is needed, since each input sample is a variable-length sequence of TRIBEv2 brain responses defined on a fixed cortical mesh. At each time step, the response is represented as a vertex-wise activation pattern with $V = 20,484$ dimensions. Directly conditioning the inference network on these full vertex-level sequences would require learning from very high-dimensional inputs relative to the available simulation budgets and the six-dimensional target space.

The summary backbone first applies a fixed, non-learnable spatial projection obtained by principal component analysis (PCA) on the training responses. This projection maps vertex-level activation into a lower-dimensional basis capturing dominant spatial response patterns. The projected features are then processed by a small residual MLP and aggregated across time using masked mean pooling, yielding the final vector representation $h(\X)$.~The summary backbone configuration is described in \autoref{app:architecture}.

\paragraph{Flow matching} Conditioned on the summary representation $h(\X)$, we model the posterior distribution over the six latent headline parameters using a conditional flow-matching inference network \citep{lipman2023flow,wildberger2023flow}. The network learns a time-dependent velocity field $\hat{u}(\thetab_t, h(\X), t)$ that transports samples from a simple Gaussian base distribution to the posterior distribution $q(\thetab \mid h(\X))$. 
Specifically, for each training pair $(\thetab, \X)$, we sample $\epsilonb \sim \mathcal{N}(\mathbf{0}, \mathbf{I})$ and $t \sim \mathcal{U}(0,1)$, and define the linear interpolation $\thetab_t = (1-t)\thetab + t\epsilonb$.
The target velocity along this path is $v(\thetab_t,t) = \epsilonb - \thetab$, which points from the latent parameter vector $\thetab$ toward the Gaussian noise sample $\epsilonb$. Thus, we minimize the conditional flow-matching objective
\begin{equation}
\min_{u, h}\,
\mathbb{E}_{p(\thetab, \X, \thetab_t, t)}
\left[
    \omega_t
    \left\Vert
    u(\thetab_t, h(\X), t) - v(\thetab_t,t)
    \right\Vert_2^2
\right].
\end{equation}
At inference time, we sample from $q(\thetab \mid h(\X))$ by first drawing
$\thetab_1 \sim \mathcal{N}(\mathbf{0}, \mathbf{I})$ and then integrating the learned reverse-time ordinary differential equation from $t=1$ to $t=0$,
\begin{equation}
    \frac{d\thetab_t}{dt} = \hat{u}(\thetab_t, h(\X), t),
\end{equation}
using a numerical ODE solver. The resulting endpoint $\thetab_0$ is taken as a posterior sample of the latent headline parameters $\thetab$ (see \autoref{app:architecture} for the hyperparameters of $\hat{u}$).

\paragraph{Simulation-based training} 

Because the TRIBEv2 responses vary in temporal length, sequences are padded or truncated to a common training length $T_{\mathrm{max}} = 48$, which affects only a small number of unusually long sequences.
Training then minimizes a flow-matching objective for posterior inference over the six latent headline-control variables (see \autoref{app:architecture}).
All networks, training procedures, and diagnostics are implemented using the \texttt{BayesFlow} library \citep{kuhmichel2026bayesflow}.

\begin{figure*}[t]
    \centering
    \includegraphics[width=.99\linewidth]{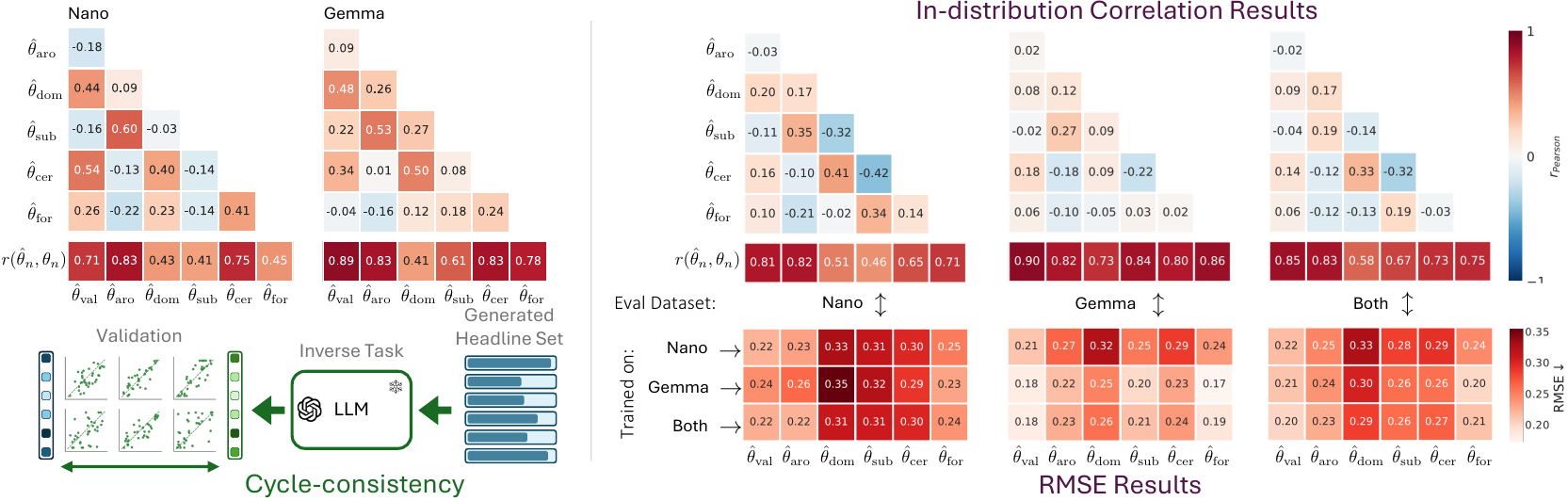}
    \caption{\textit{Cycle consistency (left) and Bayesian inversion results (right)}. \textbf{Left panel}: Cross-correlation matrices of the LLM-recovered scores show that inverse prompting introduces spurious correlations in the estimates between some dimensions (e.g., dominance with valence). Correlations between the true and recovered scores (bottom row) indicate that the generated texts retain information about the original scores, although recovery consistency differs across the two LLMs. \textbf{Right panel}: Simulation-based inference on brain maps indicates fewer spurious correlations and better score recovery from emulated brain responses. Correlation and Bayes RMSE are shown for within-dataset and cross-dataset evaluations using GPT Nano 5.4-generated, Gemma4 E4B-generated, and mixed headline datasets.}
    \label{fig:cc-results}
\end{figure*}

\subsection{Evaluation metrics}

\paragraph{Evaluating cycle consistency}
For the inverse prompt task, we evaluate the recovered scores \smash{$\hat{\thetab}^{\mathrm{text}}_n$} using two metrics. First, we estimate the Pearson correlation between each recovered parameter and its ground-truth counterpart across the validation set to verify if text-level encoding of latent scores is successful. Second, we estimate the Pearson cross-correlation matrix across recovered dimensions. 

Since the latent score dimensions are sampled independently under the prior, and the inverse prompting scheme instructs each dimension to be estimated from unique textual cues rather than from typical co-occurrence patterns, we expect low off-diagonal cross-correlations in the recovered parameters. Strong off-diagonal structure would instead indicate residual entanglement between dimensions in the generation or recovery process.

\paragraph{Evaluating simulation-based inference}
To evaluate the brain-to-score model, we first reuse the same correlation-based diagnostics, but apply them to the posterior mean estimate
\begin{equation}
    \bar{\thetab}_n = \frac{1}{S} \sum_{s=1}^{S} \thetab_n^{(s)}, 
    \qquad \thetab_n^{(s)} \sim q(\thetab \mid \X_n),
\end{equation}
where $S$ denotes the number of posterior samples. We then estimate the per-parameter Pearson correlation with ground truth and the cross-correlation matrix across recovered dimensions based on $\bar{\thetab}_n$.

As in the cycle-consistency setting, the desired behavior is weak off-diagonal cross-correlation. Whereas cycle consistency tests recoverability from text alone, high correlation with ground truth here shows both successful encoding of the latent linguistic properties in TRIBEv2 brain responses and successful recovery of these properties by the SBI model.

Further, we report per-parameter Bayes root-mean-square error (RMSE), averaged over the validation split as a complementary measure of estimation error that accounts for posterior precision and variance. 
We also inspect calibration and recovery plots as supplementary posterior diagnostics as part of an amortized Bayesian workflow \citep{kuhmichel2026bayesflow}; these diagnostics are reported in \autoref{app:experimental-details}.

\section{Related work}

\paragraph{Simulation-based inference in neuroscience}
SBI has been applied across all scales of neural modeling. At the \textit{microscopic level}, it has been repeatedly utilized for parameter identification in mechanistic single-cell models \citep{gonccalves2020training, lueckmann2017flexible, beck2022efficient}. At the \textit{mesoscale}, tools like the Human Neocortical Neurosolver (HNN) rely on SBI to link localized circuit dynamics with EEG/MEG observables \citep{neymotin2020human, jas2023hnn}. Finally, at the \textit{macroscopic level}, SBI has been used to infer connectome wiring rules \citep{boelts2023simulation}, whole-brain spectral properties \citep{bernardo2024simulation}, and large-scale network parameters \citep{hashemi2024simulation}. Our work extends this trajectory as the first to employ SBI for decoding from macroscale predictive brain models.

\paragraph{Neural decoding and stimulus reconstruction}

Beyond parameter estimation, a parallel line of research focuses on \textit{neural decoding}—the reconstruction of external stimuli from brain activity. Notable successes include image synthesis from fMRI using deep generative models \citep{scotti2023reconstructing, shen2019deep} and the decoding of continuous semantic language or speech from both invasive and non-invasive recordings \citep{tang2023semantic, willett2023high}. While deep learning has become the standard for neural decoding \citep{glaser2020machine}, most decoders are trained directly on empirical data rather than coupled to generative emulators (i.e., encoders). A recent exception is BrainCoDec \citep{nan2026meta} which enables in-context inversion. We contribute to this very recent line of work by showing that SBI can streamline the encoding-decoding loop enabled by emerging foundation brain emulators.

\section{Experiments}

\subsection{Cycle consistency with inverse prompting}\label{sec:datagen-cc}

\paragraph{Setup} We begin with a text-level cycle-consistency experiment that asks whether the sampled score vector remains recoverable when encoded into a headline block alone. Additionally, the results provide a first reference point for the subsequent SBI-based recovery results.

We generate two held-out synthetic validation splits, one generated with OpenAI's GPT-5.4 Nano \citep[Nano;][]{openai2026gpt54mininano} and one with Google DeepMind's Gemma 4 E4B \citep[Gemma;][]{googledeepmind2026gemma4modelcard}, each containing 100 headline files. Both splits are balanced across the same ten topic categories: \textit{weather, sports, infrastructure, science, culture, civic life, business, hazards, technology}, and \textit{health}, with 10 files per topic. Each file is generated from a single topic--score pair $(c_n, \thetab_n)$ and contains five headlines. Individual headlines are constrained to a typical headline length of roughly 10 to 20 words.

We then apply the inverse prompting procedure from \autoref{app:data-generation} to each headline set, yielding one recovered score vector \smash{$\hat{\thetab}^{\mathrm{text}}_n$} per sample. For both validation splits, inverse recovery is performed with a fixed Nano-based evaluator, since it could be reliably constrained to return numeric scores in the required output format. 

\paragraph{Cycle consistency results} The left panel of \autoref{fig:cc-results} summarizes the resulting cycle consistency evaluation. Recovered scores show meaningful positive correlation with their ground-truth counterparts, indicating that the generated headline sets retain a substantial fraction of the intended control signal. However, several dimensions also exhibit noticeable off-diagonal cross-correlations, showing that inverse prompting does not cleanly disentangle the underlying score variables. 
Thus, the text-only inverse task confirms basic recoverability of the latent controls, while providing a limited baseline for the later SBI-based recovery results.

\subsection{Inferring latent scores from emulated brain activity maps}

\paragraph{Setup}
We next test whether latent headline-control scores can be recovered from synthetic TRIBEv2 brain responses. Using the same generation pipeline as above, we generate 1,000 headline files per topic and convert them into temporally aligned multimodal stimuli following \autoref{sec:method-data-cc}, yielding paired brain-score examples $(\X_n,\thetab_n)$. This results in topic-balanced training sets of size $10$k for Nano, $10$k for Gemma, and $20$k for the combined setting (Both). Evaluation uses the same held-out Nano and Gemma validation splits as in the cycle-consistency experiment. All three SBI models share the same architecture and differ only in their training data (see \autoref{app:architecture} for more details on setup and training).

\paragraph{Estimation results}
The right panel of \autoref{fig:cc-results} summarizes score recovery from TRIBEv2 brain maps. Across all settings, the SBI results exceed the cycle consistency check in both agreement with ground truth and cross-correlation structure. In other words, decoded posterior mean estimates match the true scores better, while recovered dimensions remain much less entangled. These results imply that cycle consistency cannot serve as a strict upper bound on SBI performance but can provide rough ballpark figures.

Three patterns stand out. First, performance is strongest in distribution: Nano-trained models perform best on Nano validation data, and Gemma-trained models perform best on Gemma validation data. Second, cross-generator transfer is weaker, indicating that Nano- and Gemma-generated stimuli induce partially different brain-encoding statistics. Third, the combined model yields the most balanced performance across splits, and its RMSE is consistently lower on Gemma than on Nano validation data, suggesting that Gemma-generated headlines produce a cleaner signal for score inference.

Overall, these results show that TRIBEv2 brain maps contain sufficient stimulus-specific information for approximate latent-score recovery, and that SBI can yield faithful posterior approximations to the emulator-defined inverse problem.

\begin{figure}[h]
    \centering
    \includegraphics[width=0.7\linewidth]{figures/real-radar.pdf}
    \caption{\textit{Real-world point estimate results}. Inferred outlet-level latent score profiles for real-world headlines on shared AI-related topics reported by Associated Press (AP), Financial Times (FT), The Guardian, and Reuters. Small differences appear in valence, subjectivity, and formality, whereas arousal, dominance, and certainty appear to be encoded similarly.}
    \label{fig:real-world-radar}
\end{figure}

\subsection{Estimating latent scores of real-world headlines}

\paragraph{Setup} Finally, we ask whether the learned latent score space transfers from synthetic to real headlines. Because no ground-truth score vectors are available, this experiment is qualitative and focuses on whether the model yields plausible latent score profiles on natural headlines.

To this end, we collected five AI-related headlines from each of four outlets: Associated Press (AP), Financial Times (FT), The Guardian, and Reuters; we focused solely on stories covered by all four. We then processed them through the same forward pipeline as in the synthetic setting. The full headline sets and extended results are shown in \autoref{fig:real-world-ext}. For estimating the posterior distributions we reused the model trained on both, Nano and Gemma, datasets.

\paragraph{Results} A summary of the resulting outlet-level score profiles are shown in \autoref{fig:real-world-radar}. Qualitatively, the inferred score differences are plausible from the headlines. The main outlet separation occurs in valence, subjectivity, and formality while arousal, dominance and certainty stay similar. 
In particular, FT is consistently more negative/less positive in framing, AP is more formal and more interpretive, and Reuters is more colloquial. This suggests that the learned latent score space captures stylistic variation across outlets while remaining relatively stable under shared topic content.

\section{Conclusion}

This study provides a proof-of-concept framework for inverting foundation models of brain function with simulation-based inference (SBI). We connected controllable LLM-based stimulus generation, TRIBEv2 brain emulation, and SBI to recover latent linguistic properties of synthetic and real news headlines from synthetic brain maps. This was possible with surprisingly small simulation budgets.

Our study has several limitations. Our experiments are restricted to synthetic brain responses from a single foundation model, a small set of linguistic control dimensions, and headline-style stimuli. The inferred posteriors therefore validate information encoded by the emulator, not direct measurements of human brain activity. In addition, the performance of LLM-based stimulus generators depends on carefully crafted prompts.

Despite these limitations, the results point toward a broader utility of inverting foundation brain models. SBI could be used to design stimuli that target specific neural states, compare competing theories of representation, or search over text, audio, image, and video spaces for stimuli with desired neural effects. In applied settings, the same idea can lead to simulation-based tools in neuromarketing, audience modeling, and human-centered design: rather than only asking how people might respond to a stimulus, one could optimize stimuli against predicted neural and behavioral signatures before running costly empirical studies.


\section*{Acknowledgements}
NB and STR are supported by the National Science Foundation under Grant No.~2448380.

\bibliography{references}
\bibliographystyle{icml2026}

\newpage
\appendix
\onecolumn

\section{Data generation}\label{app:data-generation}

\subsection{Headline generation}
The forward generation step conditions the language model on a six-dimensional latent score vector and a topic. The system prompt defines the mechanical mapping from latent dimensions to linguistic properties. The user prompt provides the topic, numeric scores sampled from uniform priors, verbal bins, and output constraints. The same prompt logic is used for both Nano and Gemma. Training and validation sets were generated from separate latent draws, and exact string matching confirmed that no generated headline was duplicated across the retained splits.

The system prompt reads:
\begin{promptbox}
You are a linguistic engineer generating synthetic headlines for a research task.

Task:
Encode 6 independent dimensions into a headline using the specific structural rules below. 

Rules (The Mechanical Mapping):
1. Valence (Sentiment): 
   - Low (0.0): Use negative-charge nouns/adjectives (Crisis, Failure, Threat).
   - High (1.0): Use positive-charge nouns/adjectives (Benefit, Success, Growth).
2. Arousal (Energy):
   - Low (0.0): Use static, low-impact verbs (Remains, Exists, Continues) and avoid verbs of change, escalation, or sudden events.
   - High (1.0): Use high-velocity, high-impact verbs (Blasts, Surges, Explodes, Slams, Sparks). 
3. Dominance (Agency): 
   - Low (0.0): Use passive voice / patient-as-subject (e.g., "System is hit by...") and avoid naming or implying a controlling agent.
   - High (1.0): Use active voice / agent-as-subject (e.g., "Agency Enforces...").
4. Subjectivity (Interpretation): 
   - Low (0.0): Use concrete nouns and numbers; no evaluative adjectives.
   - High (1.0): Use heavy modifiers and interpretive framing (e.g., "Shocking," "Stunning").
5. Certainty (Modality): 
   - Low (0.0): Use epistemic hedges (e.g., Might, Could, Possible, Rumored).
   - High (1.0): Use absolute indicatives (e.g., Is, Will, Must, Confirmed).
6. Formality (Register): 
   - Low (0.0): Use monosyllabic, common Germanic words (e.g., Fix, Whack, Way) and clearly everyday, non-institutional phrasing.
   - High (1.0): Use multisyllabic, technical Latinate words (e.g., Implement, Methodology, Mitigation).

Orthogonality Instruction:
These 6 knobs are independent. You can have High Formality with Low Dominance ("Administrative collapse is observed") or High Valence with Low Certainty ("Potential for joy remains possible").

Output format:
- Return exactly the requested number of headlines.
- One headline per line.
- No numbering.
- No bullets.
- No prefatory text.
- No trailing commentary.
\end{promptbox}

And the forward user prompt reads:
\begin{promptbox}
Topic: {topic}

Target scores:
- Valence: {valence:.2f} ({valence_bin})
- Arousal: {arousal:.2f} ({arousal_bin})
- Dominance: {dominance:.2f} ({dominance_bin})
- Subjectivity: {subjectivity:.2f} ({subjectivity_bin})
- Certainty: {certainty:.2f} ({certainty_bin})
- Formality: {formality:.2f} ({formality_bin})

Constraints:
- Each headline should be between {min_words} and {max_words} words.
- Return exactly {num_headlines} headlines.
- One headline per line.
\end{promptbox}

\subsection{Cycle consistency}
The inverse evaluator receives the generated headline block and recovers one shared score per latent dimension for the set as a whole.

The system prompt for the inverse task reads:
\begin{promptbox}
You are a linguistic analyst decoding synthetic headlines. 

Task:
Recover the underlying score [0.0 to 1.0] for each dimension by identifying specific structural markers.

The 6-Dimension Decoding Key:
1. Valence (Sentiment): 
   - Low (0.0): Uses negative-charge nouns/adjectives (Crisis, Failure, Threat).
   - High (1.0): Uses positive-charge nouns/adjectives (Benefit, Success, Growth).
2. Arousal (Energy):
   - Low (0.0): Uses static, low-impact verbs (Remains, Exists, Continues).
   - High (1.0): Uses high-velocity, high-impact verbs (Blasts, Surges, Explodes, Slams, Sparks). 
3. Dominance (Agency): 
   - Low (0.0): Uses passive voice / patient-as-subject (e.g., "System is hit by...").
   - High (1.0): Uses active voice / agent-as-subject (e.g., "Agency Enforces...").
4. Subjectivity (Interpretation): 
   - Low (0.0): Uses concrete nouns and numbers; no evaluative adjectives.
   - High (1.0): Uses heavy modifiers and interpretive framing (e.g., "Shocking," "Stunning").
5. Certainty (Modality): 
   - Low (0.0): Uses epistemic hedges (e.g., Might, Could, Possible, Rumored).
   - High (1.0): Uses absolute indicatives (e.g., Is, Will, Must, Confirmed).
6. Formality (Register): 
   - Low (0.0): Uses monosyllabic, common Germanic words (e.g., Fix, Whack, Way) and clearly everyday, non-institutional phrasing.
   - High (1.0): Uses multisyllabic, technical Latinate words (e.g., Implement, Methodology, Mitigation).

Scoring Strategy:
- Infer one shared score per dimension for the headline set as a whole.
- Base each score on recurring observable cues in wording, framing, and structure.
- Do not interpret the absence of high-intensity markers as moderate intensity; if only low-intensity markers are present, treat the dimension as low.
- Score each dimension from the text itself.
- Do not rely on assumptions about how headline properties typically co-occur.
- Return your best estimate even if the evidence is mixed.
- Do not explain your reasoning.
\end{promptbox}

And the user prompt template for providing the headline block:
\begin{promptbox}
Estimate the stylistic scores for the following headlines.

Headlines:
{headline_block}

Return one score per dimension.
\end{promptbox}

With Nano we were able to restrict the response to the following format:
\begin{promptbox}
{
    "type": "json_schema",
    "name": "recovered_scores",
    "schema": {
        "type": "object",
        "properties": {
            "valence": {"type": "number", "minimum": 0.0, "maximum": 1.0},
            "arousal": {"type": "number", "minimum": 0.0, "maximum": 1.0},
            "dominance": {"type": "number", "minimum": 0.0, "maximum": 1.0},
            "subjectivity": {"type": "number", "minimum": 0.0, "maximum": 1.0},
            "certainty": {"type": "number", "minimum": 0.0, "maximum": 1.0},
            "formality": {"type": "number", "minimum": 0.0, "maximum": 1.0},
        },
        "required": [
            "valence",
            "arousal",
            "dominance",
            "subjectivity",
            "certainty",
            "formality",
        ],
        "additionalProperties": False,
    },
    "strict": True,
}
\end{promptbox}

\subsection{Stimulus and emulator preprocessing}
We loaded the public cortical TRIBEv2 checkpoint. Headlines were synthesized with the Piper \texttt{en\_US-lessac-medium} English voice. Word-level timings were obtained using WhisperX \texttt{large-v3} with English alignment, float16 inference, and batch size 16. The resulting text, audio, and aligned word events were passed to TRIBEv2 to generate synthetic average-subject responses on the fsaverage5 cortical mesh with 20,484 vertices.
We retained TRIBEv2’s default five-second temporal offset for compensation of the hemodynamic response delay.

\section{Architectures details}\label{app:architecture}

\subsection{Summary network}
The summary network maps a masked TRIBEv2 response tensor $\X \in \mathbb{R}^{T_{\max} \times 20484}$ to a summary embedding $\z \in \mathbb{R}^{128}$, where sequences are padded or truncated to $T_{\max}=48$ and a corresponding mask indicates valid frames. 
Responses are first standardized vertex-wise using fixed training-set means and standard deviations, and are then projected onto a fixed PCA basis with $k=4096$ components. We use this non-trainable down-projection to reduce the spatial dimensionality before learning and to avoid introducing a very large trainable projection layer in a regime with only $\sim 10$k training examples. 
The projected features are then passed through a residual MLP block that maps $(T_{\max},4096)$ to $(T_{\max},128)$ and consists of $\mathrm{LayerNorm} \rightarrow \mathrm{Dense}(128,\mathrm{GELU}) \rightarrow \mathrm{Dropout}(0.1) \rightarrow \mathrm{Dense}(128) \rightarrow \mathrm{Dropout}(0.1)$, together with a linear skip projection from the input to the output width. 
The resulting frame-wise features are aggregated by masked mean pooling over time. We use masked mean pooling because frame-to-frame correlations were very high (around 0.95 on average), and more complex temporal modules such as 1D convolutions or attention-based pooling did not improve over this simple baseline. 
Finally, a summary head of the form $\mathrm{LayerNorm} \rightarrow \mathrm{Dense}(128,\mathrm{GELU}) \rightarrow \mathrm{Dropout}(0.1) \rightarrow \mathrm{Dense}(128)$ maps the pooled representation to the final summary vector $\z$, which is passed to the downstream inference network.

\subsection{Inference network}
For posterior inference, we used BayesFlow's conditional flow-matching \citep{lipman2023flow} estimator with a \texttt{TimeMLP} backbone \citep{kuhmichel2026bayesflow}. 
Flow matching defines a forward interpolation between base noise $\thetab_1 \sim p_1$ and the target parameter vector $\thetab_0 \sim p(\thetab \mid \z)$, and trains a neural velocity field $\hat{u}(\thetab_t, t, \z)$ to approximate the conditional transport velocity along this path. At inference time, posterior samples are obtained by sampling $\thetab_1 \sim p_1$ and numerically integrating the learned velocity field from $t=1$ to $t=0$ \citep[for an overview, see][]{arruda2025diffusion}. 
The backbone is a conditional MLP that receives the current flow state, the conditioning summary $\z$, and a scalar flow time. The time input is encoded by a sinusoidal time-embedding layer and injected into the MLP blocks. We used two hidden layers with width 128, a time-embedding dimension of 8, and dropout rate 0.05. All other inference-network settings were kept at the BayesFlow defaults.

\subsection{Training details}
All models were trained with AdamW using learning rate $10^{-4}$ and weight decay $10^{-4}$. We used a batch size of $32$ and trained for at most $70$ epochs. In practice, most runs stopped after roughly $30$-$40$ epochs due to early stopping. 
During training, we monitored validation loss throughout. The model weights from the best validation epoch were retained, training was stopped if validation loss did not improve for $10$ consecutive epochs, and the learning rate was halved after $4$ consecutive epochs without improvement, down to a minimum of $10^{-6}$. All training runs were performed on a single NVIDIA GeForce RTX 4090 GPU and took approximately one hour.

\section{Experimental details}\label{app:experimental-details}

\subsection{Extended SBI results}
\begin{figure}
    \centering
    \includegraphics[width=0.99\linewidth]{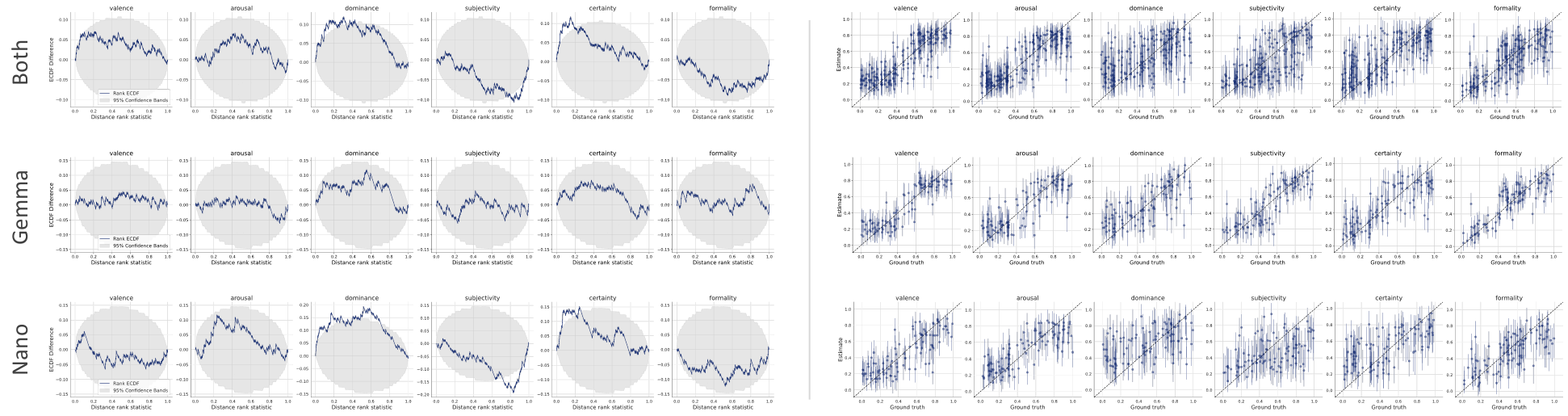}
    \caption{\textit{Extended in-distribution SBI diagnostics.} Rows correspond to the combined model, the Gemma-only model, and the Nano-only model. \textbf{Left:} ECDF calibration diagnostics for the six latent score dimensions. Blue curves show rank-ECDF differences and grey regions indicate $95\%$ confidence bands. \textbf{Right:} recovery plots showing posterior samples and uncertainty intervals against the known generating parameters. Quantitative recovery correlations and RMSE values are reported in \autoref{fig:cc-results}. All diagnostics use $N=500$ posterior samples per validation instance.}
    \label{fig:recovery-ecdf}
\end{figure}

We report additional in-distribution recovery and calibration diagnostics in \autoref{fig:recovery-ecdf}. Each model was evaluated on validation data from the same headline generator used during training: Gemma on Gemma, Nano on Nano, and the combined model on the combined validation set. For each validation instance, we drew $N=500$ posterior samples. The Gemma and Nano evaluations each contain $100$ validation instances, while the combined evaluation contains $200$ instances.

The quantitative recovery correlations and RMSE values are reported in \autoref{fig:cc-results}. The appendix recovery plots instead visualize the posterior samples and uncertainty intervals underlying these summaries. Across all in-distribution settings, posterior estimates show a clear positive association with the known generating scores. 
The recovery plots also reveal two broad regions of validation points, corresponding to low- and high-score regimes induced by the controlled headline-generation procedure, rather than a uniformly filled parameter range. 
Although the six controls are sampled independently, their LLM-generated linguistic realizations are not guaranteed to be orthogonal: shared lexical, syntactic, and affective cues can induce generator-specific dependencies and unequal identifiability across dimensions. The marker-based prompts and verbal score bins may also contribute to the regime-like structure visible in \autoref{fig:recovery-ecdf}.
Recovery is most stable for valence, arousal, and formality, while dominance is consistently noisier and less tightly identified. The Nano model exhibits visibly wider posterior intervals and stronger scatter, whereas the Gemma model shows the cleanest posterior recovery structure.

Calibration was assessed using BayesFlow ECDF diagnostics. The rank-ECDF curves remain largely within the $95\%$ confidence bands, indicating no severe calibration failure. At the same time, several dimensions show mild structured deviations from the expected ECDF behavior, especially for the Nano and combined models. Overall, the diagnostics support reliable in-distribution posterior recovery with approximate, but not perfectly uniform, calibration. Taken together, valence, arousal, and formality appear sufficiently stable for interpretation in the in-distribution analyses, with subjectivity and certainty usable but somewhat less robust. Dominance should be interpreted more cautiously, since it shows weaker recovery and broader posterior uncertainty across models.

\subsection{Real-world headlines}

\begin{figure}
    \centering
    \includegraphics[width=0.99\linewidth]{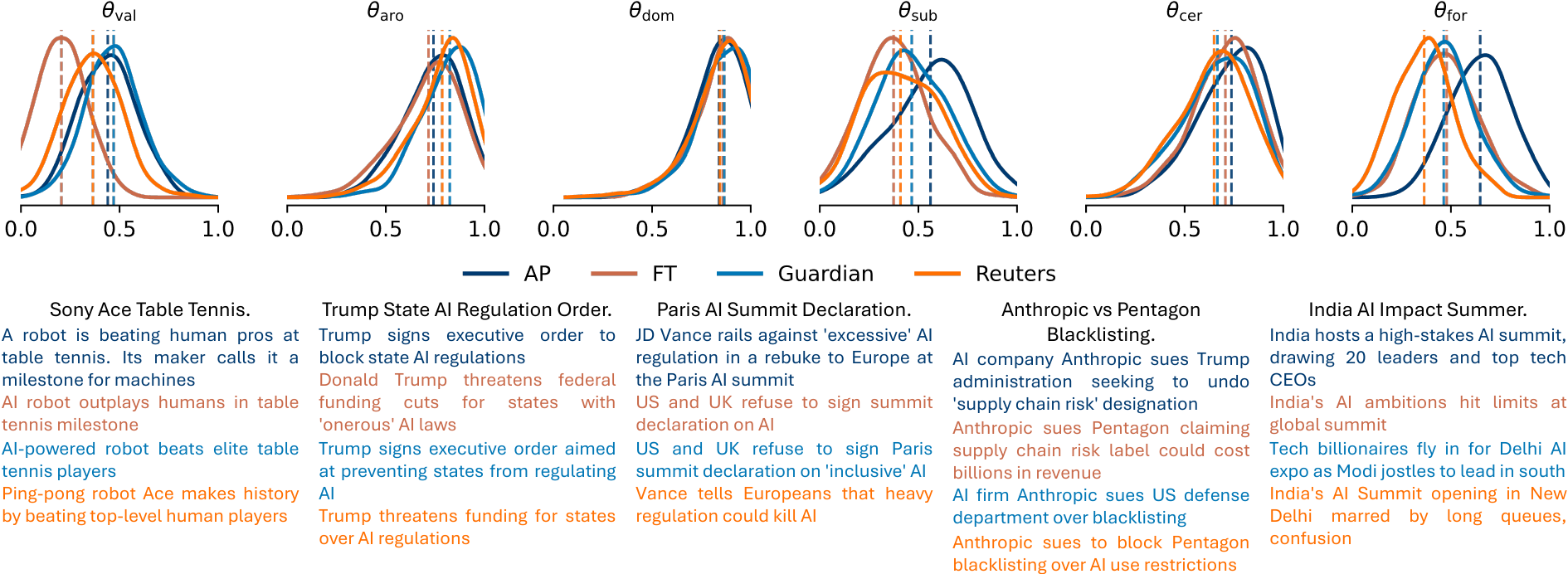}
    \caption{\textit{Detailed qualitative results for real-world AI headlines.} \textbf{Top:} kernel density estimates of the inferred latent score posterior distributions for Associated Press (AP), Financial Times (FT), The Guardian, and Reuters across the six score dimensions; dashed vertical lines mark the corresponding posterior means. \textbf{Bottom:} matched headline sets used for inference, grouped by shared news event and color-coded by outlet. The underlying headline sources are listed in \autoref{tab:real-world-headlines}. The strongest outlet differences appear in valence, subjectivity, and formality: FT is inferred as less positive in framing, AP as more formal and interpretive, and Reuters as more colloquial, while arousal, dominance, and certainty remain comparatively similar across outlets.}
    \label{fig:real-world-ext}
\end{figure}

We additionally evaluated the trained inverse model on real-world AI headlines. To enable outlet-level comparisons, we selected five shared news events that were each covered by AP, Financial Times (FT), The Guardian, and Reuters. For each event, one headline per outlet was used as the stimulus, yielding matched headline sets across outlets. The inferred posterior distributions and the corresponding headline sets are shown in \autoref{fig:real-world-ext}. The headline sources are listed in \autoref{tab:real-world-headlines}.

\begin{table*}[t]
    \centering
    \scriptsize
    \setlength{\tabcolsep}{3pt}
    \renewcommand{\arraystretch}{1.18}
    \caption{\textit{Real-world headline sources.} Each table entry reports the publication date and source link for the corresponding outlet-event pair. Rows correspond to outlets and columns to the five matched AI-related news events used in the real-world posterior inference experiment. The corresponding headline texts are shown in \autoref{fig:real-world-ext}.}
    \label{tab:real-world-headlines}
    \begin{tabularx}{\linewidth}{lYYYYY}
        \toprule
        \textbf{Outlet}
        & \textbf{Sony Ace table tennis}
        & \textbf{Trump state AI regulation}
        & \textbf{Paris AI summit}
        & \textbf{Anthropic vs Pentagon}
        & \textbf{India AI summit} \\
        \midrule

        AP
        & 2026-04-22, \href{https://apnews.com/article/ai-table-tennis-robot-ping-pong-sony-995b239945e0dc8d7bea918a850969dc}{source}
        & 2025-12-11, \href{https://apnews.com/article/trump-ai-regulation-executive-order-state-laws-9cb4dd1bc249e404260b3dc233217388}{source}
        & 2025-02-11, \href{https://apnews.com/article/1d7826affdcdb76c580c0558af8d68d2}{source}
        & 2026-03-09, \href{https://apnews.com/article/anthropic-trump-pentagon-hegseth-ai-104c6c39306f1adeea3b637d2c1c601b}{source}
        & 2026-02-16, \href{https://apnews.com/article/india-ai-modi-macron-lula-f32aa15263349dadad1840dadc75ca83}{source} \\

        FT
        & 2026-04-22, \href{https://www.ft.com/content/9860f042-3332-4534-9b1a-fa9f57b8347e}{source}
        & 2025-12-11, \href{https://www.ft.com/content/a114351c-2f4f-4688-96d9-5ae1af777420}{source}
        & 2025-02-11, \href{https://www.ft.com/content/a6b5426d-645f-433b-8090-a2a26a3deec6}{source}
        & 2026-03-09, \href{https://www.ft.com/content/af404e0a-7abc-49bc-9584-cd4690152f86}{source}
        & 2026-02-22, \href{https://www.ft.com/content/5c26f2f6-c857-407c-93fe-7f59aa88c8f4}{source} \\

        The Guardian
        & 2026-04-22, \href{https://www.theguardian.com/science/2026/apr/22/ai-powered-robot-beats-elite-table-tennis-players-milestone-robotics}{source}
        & 2025-12-11, \href{https://www.theguardian.com/us-news/2025/dec/11/trump-executive-order-artificial-intelligence}{source}
        & 2025-02-11, \href{https://www.theguardian.com/technology/2025/feb/11/us-uk-paris-ai-summit-artificial-intelligence-declaration}{source}
        & 2026-03-09, \href{https://www.theguardian.com/technology/2026/mar/09/anthropic-defense-department-lawsuit-ai}{source}
        & 2026-02-18, \href{https://www.theguardian.com/technology/2026/feb/18/delhi-ai-expo-modi-jostles-lead-south}{source} \\

        Reuters
        & 2026-04-22, \href{https://www.reuters.com/sports/ping-pong-robot-ace-makes-history-by-beating-top-level-human-players-2026-04-22/}{source}
        & 2025-12-11, \href{https://www.reuters.com/world/trump-says-he-will-sign-order-curbing-state-ai-laws-2025-12-11/}{source}
        & 2025-02-11, \href{https://www.reuters.com/technology/us-britain-have-not-signed-paris-ai-summit-declaration-2025-02-11/}{source}
        & 2026-03-09, \href{https://www.reuters.com/world/anthropic-sues-block-pentagon-blacklisting-over-ai-use-restrictions-2026-03-09/}{source}
        & 2026-02-16, \href{https://www.reuters.com/world/india/indias-ai-summit-opening-new-delhi-marred-by-long-queues-confusion-2026-02-16/}{source} \\

        \bottomrule
    \end{tabularx}
\end{table*}

\section{Code availability}
The source code is publicly available under \href{https://github.com/stefanradev93/brain-inversion}{https://github.com/stefanradev93/brain-inversion}.

\end{document}